\documentclass[11pt,a4paper]{article}
\usepackage[hyperref]{emnlp2020}
\usepackage{times}
\usepackage{latexsym}

\usepackage{microtype}

\aclfinalcopy 
\def\aclpaperid{2460} 
\usepackage{graphicx}
\usepackage{amsmath}
\usepackage{amsfonts}
\usepackage{amsthm}
\usepackage{booktabs}
\usepackage{algorithm}
\usepackage{algorithmic}
\usepackage{array}
\usepackage{float}
\usepackage{booktabs}
\usepackage{times}
\usepackage{soul}
\usepackage{url}
\usepackage{color}
\title{A Sentiment-Controllable Topic-to-Essay Generator \\ with Topic Knowledge Graph}

\author{
    Lin Qiao\textsuperscript{1}\Thanks{This work is done when Lin Qiao was interning at Pattern Recognition Center, WeChat AI, Tencent Inc, China}, Jianhao Yan\textsuperscript{2},  Fandong Meng\textsuperscript{2}, Zhendong Yang, and Jie Zhou\textsuperscript{2}\\
    \textsuperscript{1}School of Software and Microelectronics, Peking University\\
    \textsuperscript{2}Pattern Recognition Center, WeChat AI, Tencent Inc, Beijing, China\\
    qiaolin66666@gmail.com\\
    \{elliottyan,fandongmeng,withtomzhou\}@tencent.com
}
\date{}

\begin{document}
\maketitle
% \setlength{\abovedisplayskip}{5pt}
% \setlength{\belowdisplayskip}{5pt}
%当前摘要第一句话保留，接下来要写四句话：
%要解决的问题是什么，或者说前人方法有什么问题。
%因此我们做了什么事情，或者我们提出了什么方法。
%我们的方法大概是怎么做的。
%汇报实验效果。

\begin{abstract}
% 一个。。。的文章，需要情感。。tga的帮助
Generating a vivid, novel, and diverse essay with only several given topic words is a challenging task of natural language generation. In previous work, there are two problems left unsolved: neglect of sentiment beneath the text and insufficient utilization of topic-related knowledge.
% but remains a challenging problem 
% due to the absence of sentiment controlled and diversity driven modeling, as well as insufficient utilizing of topic-related knowledge in previous work. 
Therefore, we propose a novel \textbf{S}entiment-\textbf{C}ontrollable topic-to-essay generator with a \textbf{T}opic \textbf{K}nowledge \textbf{G}raph enhanced decoder, named SCTKG, which is based on the conditional variational auto-encoder (CVAE) framework. 
We firstly inject the sentiment information into the generator for controlling sentiment for each sentence, which leads to various generated essays. Then we design a Topic Knowledge Graph enhanced decoder. Unlike existing models that use knowledge entities separately, our model treats knowledge graph as a whole and encodes more structured, connected semantic information in the graph to generate a more relevant essay.
Experimental results show that our SCTKG can generate sentiment controllable essays and outperform the state-of-the-art approach in terms of topic relevance, fluency, and diversity on both automatic and human evaluation.
%Generating an essay with only several given topic words is a promising but challenging task of natural language generation. Being able to express rich emotions is a distinguishing characteristic of human writing. However, previous work neglected to endow generated texts with specific sentiment. Such defect hurts the vividness and diversity of the essay. Meanwhile, previous work fails to comprehensively model the topic-related knowledge graph, which hurts the expression of topic information. To address these problems, we present a novel \textbf{S}entiment \textbf{C}ontrol with \textbf{T}opic \textbf{G}raph enhanced architecture based on conditional variational auto-encoder, named SCTKG. We encode the sentiment label into the encoder and decoder of our generator to guide the sentiment of the sentence. We also present a novel Topic Graph Attention (TGA) method to learning the representation of the relationship in the topic-related knowledge. Experimental results show that our model outperforms the state-of-the-art methods in all evaluation metrics and achieves satisfactory sentiment control accuracy. 
\end{abstract}

\section{Introduction}
\textbf{T}opic-to-\textbf{e}ssay \textbf{g}eneration (TEG) task aims at generating human-like paragraph-level texts with only several given topics. It has plenty of practical applications, e.g., automatic advertisement generation, intelligent education, or assisting in keyword-based news writing \cite{leppanen2017data}. Because of its great potential in practical use and scientific research, TEG has attracted a lot of interest. \cite{feng2018topic,yang2019enhancing}. However, In TEG, two problems are left to be solved: the neglect of sentiment beneath the text and the insufficient utilization of topic-related knowledge.

\begin{figure}
\centering
\includegraphics[width=7.5cm,height=6cm]{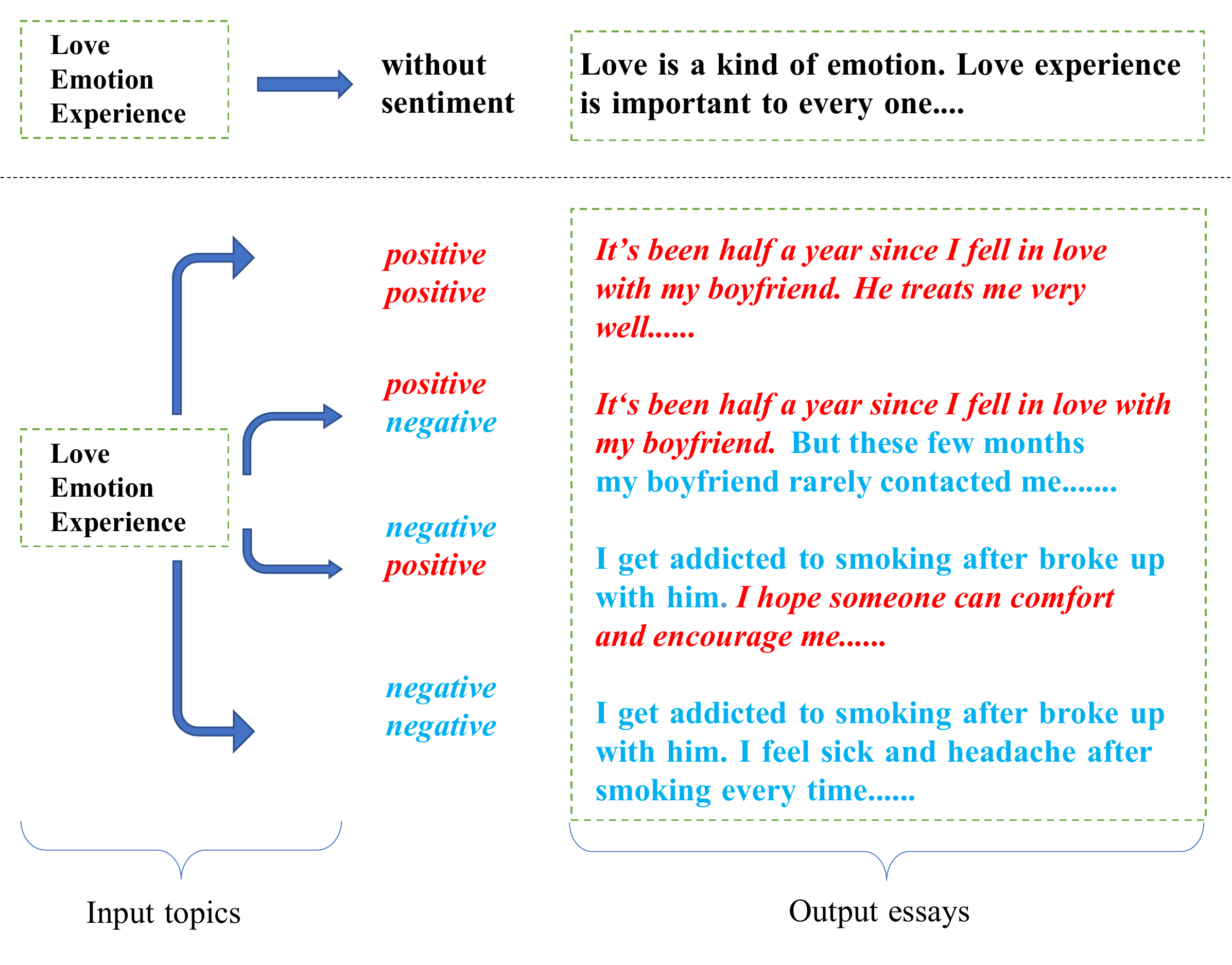}
\caption{Examples of comparison between the generated essays with sentiment control and without sentiment. We show the first two sentences for each generated essay and denote positive sentences in red and negative sentences in blue. Sentences without sentiment label are showed in black.}
\label{figure 1}
\end{figure}
A well-performed essay generator should be able to generate multiple vivid and diverse essays when given the topic words. However, previous work tends to generate dull and generic texts. One of the reason is that they neglect the sentiment factor of the text. By modeling and controlling the sentiment of generated sentences, we can generate much more diverse and fascinating essays.
% (senti 对 div 有帮助，一个bonus）
% for several topic words, there exist many proper corresponding articles. But previous work fails to generate discourse-level(突然） diverse texts and tends to generate dull and generic articles. By controlling the sentiment of each generated sentence, we can generate much more diverse and fascinating texts.
As shown in Figure~\ref{figure 1}, given the topic words ``Love'', ``Experience'' and ``Emotion'', the ``without sentiment'' model generates monotonous article. In contrast, 
% 没用senti的时候。。
% 作为对比， 我们的模型。。。 ）
the sentiment-attach model generates positive statements such as “fall in love with my
boyfriend” when given the “positive” label, and
generates negative phrases such as “addicted to
smoking”, “broke up” when given the “negative”
label. In addition, sentiment control is especially essential in topic-to-essay generation task, which aims to generate multiple sentences. As the number of sentences increases, the search space for generation model is exponentially enlarged by controlling the sentiment polarity for each of the sentence. Therefore, the ability to control sentiment is essential to improve discourse-level diversity for the TEG task.
% （对TEG任务
% 帮助，而不是seq2seq有帮助）

As for the other problem, imagine that when we human beings are asked to write articles with some topics, we heavily rely on our commonsense knowledge related to the topics. Therefore, the proper usage of knowledge plays a vital role in the topic-to-essay generation. 
% （当人类写文章的时候，我们会用到外部知识，因此teg也需要）Another challenging problem in TEG is that the input information is extremely insufficient compared to the target output.（这种不足具体会带来什么问题？） It is natural that some external knowledge should be used to alleviate the imbalance problem and （因此）how to represent and utilize external knowledge plays a vital role in our task. 
% \cite{feng2018topic} are the first to propose the TEG task and they utilize coverage vector to incorporate topic information for essay generation. However, the model performance is not satisfactory since they do not solve the above imbalance problem. 
Previous state-of-the-art method \cite{yang2019enhancing} extracts topic-related concepts from a commonsense knowledge base to enrich the input information. However, they ignore the graph structure of the knowledge base, which merely refer to the concepts in the knowledge graph and fail to consider their correlation. This limitation leads to concepts being isolated from each other. For instance, given two knowledge triples \emph{(law, antonym, disorder)} and \emph{(law, part of, theory)}, about the topic word \emph{law}, \citet{yang2019enhancing} simply uses the neighboring concepts \emph{disorder} and \emph{theory} as a supplement to the input information. However, their method fails to learn that \emph{disorder} has opposite meaning with \emph{law} while \emph{theory} is a hypernym to \emph{law}, which can be learned from their edges (correlations) in the knowledge graph. Intuitively, lacking the correlation information between concepts in the knowledge graph hinders a model from generating appropriate and informative essays.

To address these issues, we propose a novel \textbf{S}entiment-\textbf{C}ontrollable topic-to-essay generator with a \textbf{T}opic \textbf{K}nowledge \textbf{G}raph enhanced decoder, named SCTKG, which is based on the conditional variational auto-encoder (CVAE) framework. To control the sentiment of the text, we inject the sentiment information in the encoder and decoder of our model to control the sentiment from both sentence level and word level. The sentiment labels are provided by a sentiment classifier during training. To fully utilize the knowledge, the model retrieves a topic knowledge graph from a large-scale commonsense knowledge base ConceptNet \cite{speer2012representing}. Different from \citet{yang2019enhancing}, we preserve the graph structure of the knowledge base and propose a novel \emph{Topic Graph Attention} (TGA) mechanism. TGA attentively reads the knowledge graphs and makes the full use of the structured, connected semantic information from the graphs for a better generation. In the meantime, to make the generated essays more closely surround the semantics of all input topics, we adopt adversarial training based on a multi-label discriminator. The discriminator provides the reward to the generator based on the coverage of the output on the given topics.

Our contributions can be summarized as follow:
\begin{enumerate}
    \item  We propose a sentiment-controllable topic-to-essay generator based on CVAE, which can
generate high-quality essays as well as control
the sentiment. To the best of our knowledge, we are the first to control the sentiment in TEG and demonstrate the potential of our model to generate diverse essays by controlling the sentiment. 
    \item We equip our decoder with a topic knowledge graph and propose a novel Topic Graph Attention (TGA) mechanism. TGA makes the full use of the structured, connected semantic information from the topic knowledge graph to generate more appropriate and informative essays.
    \item We conduct extensive experiments, showing that our model accurately controls the sentiment and outperforms the state-of-the-art methods both in automatic and human evaluations.
\end{enumerate}

\section{Task Formulation}
Traditional TEG task takes as input a topic sequence $X$ = \(\left(x_{1}, \cdots, x_{m}\right)\) with $m$ words, and aims to generate an essay with $M$ sentences \(\left(L_{1}, \cdots, L_{M}\right)\) corresponding with topic sequence $X$. In this paper, we provide a sentiment sequence $S$ = \(\left(s_{1}, \cdots, s_{M}\right)\), each of which corresponds to a target sentence in essay. Each sentiment can be \emph{positive}, \emph{negative}, or \emph{neutral}. 

Essays are generated in a sentence-by-sentence manner. The first sentence \(L_{1}\) is generated only conditioned on the topic sequence $X$, then the model takes all the previous generated sentences as well as the topic sequence to generate the next sentence until the entire essay is completed. In this paper, we denote the previous sentences $L_{1:i-1}$ as context.
% \vspace{-0.15cm}
% \begin{align}
% L_{i}^{*}=\underset{L}{\arg \max } P\left(L | L_{1: i-1}, \mathbb{X}, S\right)  
% \end{align}

%每一段话之前说一下这是要干什么。
\section{Model Description}
\begin{figure*}
\centering
\includegraphics[width=16cm,height=5cm]{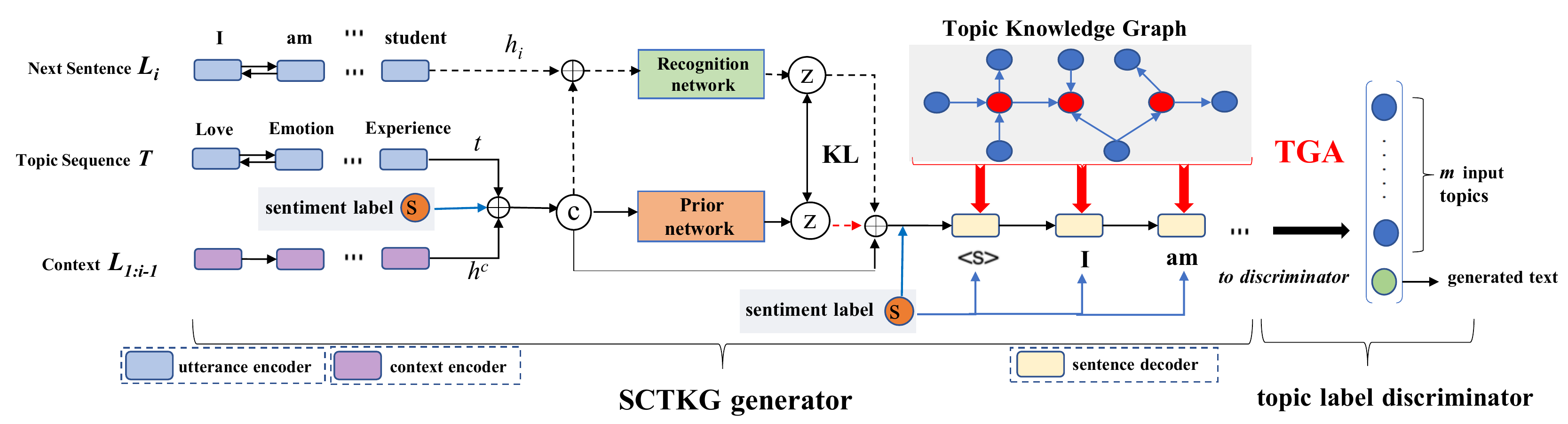}

\caption{The architecture of our model. \(\oplus\) denotes the vector concatenation operation. Only the part with solid lines and the red dotted arrow is applied at inference, while the entire CVAE except the red dotted arrow part used in the training process. Sentiment label $s$ with blue arrows denote sentiment control. Red solid lines denote TGA at each decoding step. The text generated by SCTKG generator feeds to topic label discriminator. The above \emph{m} blue circle represents the probability that it belongs to the real text with the \emph{m} input topics, and the green circle represents the given text is a generated text. }

\label{figure 3}
\end{figure*}

In this section, we describe an overview of our proposed model. Our SCTKG generator based on a CVAE architecture consists of an encoder and a topic knowledge graph enhanced decoder. The encoder encodes topic sequence, sentiment, and context and regards them as conditional variables $c$. Then a latent variable $z$ is computed from $c$ through a recognition network (during training) or prior network (during inference). The decoder attaches with a topic knowledge graph and sentiment label to generate the texts. At each decoding step, the TGA is used to enrich input topic information through effectively utilizing the topic knowledge graph.

We adopt a two-stage training approach: (1) Train the SCTKG generator with the conventional CVAE loss; (2) After the first step is done, we introduce a topic label discriminator to evaluate the performance of SCTKG generator. We adopt adversarial training to alternately train the generator and the discriminator to further enhance the performance of the SCTKG generator.
% 1.traditon topic-to-esay，问题formulation
\subsection{SCTKG Generator}
\subsubsection{Encoder}

% The utterance encoder is a bidirectional recurrent neural network (BRNN)\cite{schuster1997bidirectional}and is used to encode a single sentence and topic sequence. We concatenate the last hidden state of both forward and backward RNN to
% capture the semantic information from both sides
% , formulized as h = $
% [\overrightarrow{h_{T}}, \overleftarrow{h_{T}}]
% $where T is the sequence length. For the context encoder, we use a hierarchical encoding strategy. First, each sentence Xi in context X is encoded by the sentence encoder to get a latent representation hsenti . Then a single layer forward RNN is used to encode the sentence representations into a final state hcontext.
As shown in Figure \ref{figure 3}, the utterance encoder is a bidirectional GRU \cite{chung2014empirical} to encode an input sequence into a fixed-size vector by concatenating the last hidden states of the forward and backward GRU. We use the utterance encoder to encode the topic sequence $X$ into $h^{x}$ = $[\overrightarrow{h^{x}}, \overleftarrow{h^{x}}]$, $h^{x}$ $\in \mathbb{R}^{d}$. $d$ is the dimension of the vector. The next sequence $L_{i}$ is also encoded by utterance encoder into $h_{i}$ = $[\overrightarrow{h_{i}}, \overleftarrow{h_{i}}]$, $h_{i}$ $\in \mathbb{R}^{d}$. For context encoder, we use a hierarchical encoding strategy. Firstly, each sentence in context $L_{1:i-1}$ is encoded by utterance encoder to get a fixed-size vector. By doing so, the context $L_{1:i-1}$ is encoded into $h_{context} = [h_{1}, h_{2}\dots, h_{i-1}]$. Then a single layer forward GRU is used to encode the sentence representations  $h_{context}$ into a final state vector $h^{c} \in \mathbb{R}^{d}$.
% $$
% \begin{array}{l}
% h^{c}=\mathrm{GRU}_{c}(h_{context}) 
% \end{array}
% $$
% $h^{c} \in R$.
% [\overrightarrow{h^{x}}, \overleftarrow{h^{x}}]$. 
% Note that $L_{i}$ corresponds to $Y$ in Eq. (\ref{Y})
% % $$
% % \begin{array}{l}
% % \overrightarrow{h}_{i}=\overrightarrow{\mathrm{GRU}}_{L}\left(L_{i}\right) \\
% % \overleftarrow{h}_{i}=\overleftarrow{\mathrm{GRU}}_{L}\left(L_{i}\right)
% % \end{array}
% % $$

% % $$
% % \begin{array}{l}
% % \overrightarrow{h}^{x}=\overrightarrow{\mathrm{GRU}}_{x}\left(\mathbb{x}\right) \\
% % \overleftarrow{h}^{x}=\overleftarrow{\mathrm{GRU}}_{x}\left(\mathbb{x}\right)
% % \end{array}
% % $$
% The next sentence representation $h_{i}$ = $
% [\overrightarrow{h_{i}}, \overleftarrow{h_{i}}] $ and the topic sequence representation $h^{x}$ = $
% [\overrightarrow{h^{x}}, \overleftarrow{h^{x}}]$. 

% Context encoder uses a hierarchical encoding strategy. Firstly, each sentence in context $L_{1:i-1}$ is encoded by a biGRU to get a fixed-size vector  as follows:
% $$
% h_{context} = [h_{1}, h_{2}\dots, h_{i-1}]
% $$
% Then a single layer forward RNN is used to encode the sentence representations into a final state vector
% $$
% \begin{array}{l}
% h^{c}=\mathrm{GRU}_{c}(h_{context}) 
% \end{array}
% $$
% $h^{c} \in R$.

Then the concatenation of $h^{c}$, $h^{x}$, $e(s)$ is functionalized as the conditional vector $c$ = \(\left[e(s) ; h^{c} ; h^{x}\right]\). \(e(s)\) is the embedding of sentiment label $s$. We assume that $z$ follows a multivariate Gaussian distribution with a diagonal covariance matrix. Thus the recognition network \(q_{\phi}(z | h_{i}, c)\) and the prior
network \(p_{\theta}(z | c)\) follow  \( \mathcal{N}\left(\mu, \sigma^{2} \mathbf{I}\right)\) and \( \mathcal{N}\left(\mu^{\prime}, \sigma^{\prime 2} \mathbf{I}\right)\), respectively. \(\mathbf{I}\) is identity matrix, and then we have
\begin{align}
\begin{aligned}
\left[\mu, \sigma^{2}\right] &=\mathrm{MLP}_{\text {recognition }}(h_{i}, c), \\
\left[\mu^{\prime}, \sigma^{\prime 2}\right] &=\mathrm{MLP}_{\text {prior }}(c).
\end{aligned}
\end{align}

Additionally, we use a reparametrization trick \cite{kingma2013auto} to sample $z$ from the recognition network during training and  from prior network during testing.
\subsubsection{Decoder}
A general Seq2seq model may tend to emit generic and meaningless sentences. To create more meaningful essays, we propose a topic knowledge graph enhanced decoder. The decoder is based on a 1-layer GRU network with initial state $d_{0}$ = $W_{d}[z, c, e(s)]+b_{d}$. $W_{d}$ and $b_{d}$ are trainable decoder parameters and $e(s)$ is the sentiment embedding as mentioned above. 
% The decoder then predicts the words and the input at every decoding step is $[e(L_{i}^{t}), e(s), g_{t}]$ where $e(L_{i}^{t})$ is the embedding of $t^{th}$ word in $L_{i}$. $g_{t}$ is the graph vector get from the topic graph which will introduce later. 
% The variational lower bound to the
% loss $-log p(Y |c)$ can be expressed as:
% \begin{align}\label{loss}
% \begin{aligned}
% &-\mathcal{L}\left(\theta_{\mathrm{D}} ; \theta_{\mathrm{P}} ; \theta_{\mathrm{R}} ; Y, c\right)=\mathcal{L}_{\mathrm{KL}}+\mathcal{L}_{\mathrm{decoder}}\\
% &=\mathrm{KL}\left(q_{\mathrm{R}}(z | Y, c) \| p_{\mathrm{P}}(z | c)\right)\\
% &-\mathbb{E}_{q_{R}(z | Y, c)}\left(\log p_{\mathrm{D}}(Y | z, c)\right)
% \end{aligned}
% \end{align}
% Here, $\theta_{\mathrm{D}}$ ; $\theta_{\mathrm{P}}$ ; $\theta_{\mathrm{R}}$ are the parameters of the decoder, prior network, and recognition network, respectively. Intuitively, the second term maximizes
% the sentence generation probability after sampling
% from the recognition network, while the first term
% minimizes the distance between prior and recognition network. Besides, we also use the annealing trick and BOW-loss \cite{zhao2017learning} to alleviate the vanishing latent variable problem in VAE training.
% (TGA部分，先检索出n个三元组，
As shown in Figure \ref{figure 3}, we equip the decoder with a topic knowledge graph to incorporate commonsense knowledge from ConceptNet\footnote{\url{https://conceptnet.io}}. ConceptNet is a semantic network which consists of triples $R = (head; rel; tail)$. The head concept $head$ has the relation $rel$ with tail concept $tail$. We use word vectors to represent head and tail concepts and learn trainable vector $\boldsymbol{r}$ for relation $rel$
, which is randomly initialized.
Each word in the topic sequence is used as a query to retrieve a subgraph from ConceptNet and the topic knowledge graph is constituted by these subgraphs. Then we use the Topic Graph Attention (TGA) mechanism to read from the topic knowledge graph at each generation step. 
% \mathbf{G} = \left\{\tau_{1}, \tau_{2}, \cdots, \tau_{N}\right\}
% Specifically, each word in topic sequence is used as a query to retrieve a one-hop graph from ConceptNet. we use the topic words as queries to retrieve some commonsense knowledge graphs G = \(\left\{g_{1}, g_{2}, \cdots, g_{N_{G}}\right\}\) from knowledge base , and each topic corresponds to a graph in G. Each graph consists of a set of triples \(g_{i}=\left\{\tau_{1}, \tau_{2}, \cdots, \tau_{N_{g_{i}}}\right\}\) and each triple (head entity, relation, tail entity) is denoted as \(\tau =  (h, r, t)\). We denote the neighbouring concept of our topic words as c, note that c can either be h or t. 
% The major difference to \cite{yang2019enhancing} lies in that our graph attention encodes more structured semantic information by considering not only all nodes in a graph but also relations between nodes.

\paragraph{Topic Graph Attention.} As previously stated, a proper usage of the external knowledge plays a vital role in our task. TGA takes as input the retrieved topic knowledge graph and a query vector \(\boldsymbol{q}\) to produce a graph vector $g_{t}$. We set \(\boldsymbol{q}\) = \(\left[d_{t-1} ; c ; z\right]\), where $d_{t-1}$ represents the decoder hidden state for $t-1$ step. At each decoding step, we calculate the correlation score between each of the triples in the graph and \(\boldsymbol{q}\). Then we use the correlation score to compute the weighted sum of all the neighboring concepts\footnote{As shown in Figure \ref{figure 3}, in the topic knowledge graph, red circles denote the topic words and blue circles denote their neighboring concepts. Since we have already encoded topic information in the encoder, the graph vector $g_{t}$ in this section mainly focuses on the neighboring concept to assist the generation.} to the topic words to form the final graph vector $g_{t}$. Neighboring concepts are entities that directly link to topic words. We formalize the computational process as follows:
% For example, we have a topic word \emph{law}, and two relevant knowledge triples \emph{(law, antonym, disorder)} and \emph{(rule, related to, law)}, the concept ``\emph{disorder}(tail concept)'' and ``\emph{rule}(head concept)'' are the neighboring concepts related to the \emph{law}. Since we have already encoded topic information in the encoder, so the graph vector $g_{t}$ in this section mainly focuses on the neighboring concept to assist the generation. Formally
$$\begin{aligned}
&g_{t}=\sum_{n=1}^{N} \alpha_{n} \boldsymbol{o}_{n}\\
&\alpha_{n}=\frac{\exp \left(\beta_{n}\right)}{\sum_{j=1}^{N} \exp \left(\beta_{j}\right)}\\
&\beta_{n}=\left\{\begin{array}{lc}
\left(\mathbf{W}_{1} \boldsymbol{q}\right)^{\top} \tanh \left(\mathbf{W}_{2} \boldsymbol{r}_{n}+\mathbf{W}_{3} \boldsymbol{o}_{n}\right) \\
\text {when} \qquad \qquad \qquad \boldsymbol{o}_{n} \in \mathcal{S}_{1} \\
\left(\mathbf{W}_{1} \boldsymbol{q}\right)^{\top} \tanh \left(\mathbf{W}_{2} \boldsymbol{r}_{n}+\mathbf{W}_{4} \boldsymbol{o}_{n}\right) \\
\text {when}  \qquad \qquad \qquad \boldsymbol{o}_{n} \in \mathcal{S}_{2}
\end{array}\right.
\end{aligned}$$
% \begin{align}
% g_{t} &=\sum_{n=1}^{N} \alpha_{n}\boldsymbol{o}_{n},\\
% \alpha_{n} &=\frac{\exp \left(\beta_{n}\right)}{\sum_{j=1}^{N} \exp \left(\beta_{j}\right)} ,\\
% \beta_{n} &=\left\{\begin{array}{ll}
% {\left(\mathbf{W}_{1}\boldsymbol{q}\right)^{\top}\tanh \left(\mathbf{W}_{2} \boldsymbol{r}_{n}+\mathbf{W}_{3} \boldsymbol{o}_{n}\right)}\\  {when \qquad \qquad \qquad \boldsymbol{o}_{n} \in \mathcal{S}_{1}}\quad, \\
% {\left(\mathbf{W}_{1}\boldsymbol{q}^\right)^{\top}\tanh \left(\mathbf{W}_{2} \boldsymbol{r}_{n}+\mathbf{W}_{4} \boldsymbol{o}_{n}\right)}\\  {when \qquad \qquad \qquad \boldsymbol{o}_{n} \in \mathcal{S}_{2}}
% \end{array}\right.
% \end{align}

where \(\boldsymbol{o}_{n}\) is the embedding of $n^{th}$ neighboring concept and $\boldsymbol{r}_{n}$ is the embedding of the relation for $n^{th}$ triple in the topic knowledge graph.
% \footnote{Note that entities in ConceptNet are common words
% (such as tree, leaf, animal), we thus use word vectors to represent $head/rel/tail$ directly, instead of using geometric embedding methods (e.g., TransE) to learn entity and relation embeddings. In this way, there is no need to bridge the representation gap between geometric embeddings and text-contextual embeddings (i.e., word vectors).}
\(\mathbf{W}_{1}, \mathbf{W}_{2}, \mathbf{W}_{3}, \mathbf{W}_{4}\) are weight matrices for query, relations, head entities and tail entities, respectively. $\mathcal{S}_{1}$ contains the neighboring concepts which being the head concepts in their triples, while $\mathcal{S}_{2}$ contains the neighboring concepts which being the tail concepts. The matching score \(\beta_{n}\) represents the correlation between the query \(\boldsymbol{q}\) and neighbouring concept $o_{n}$. Essentially, a graph vector $g_{t}$ is the weighted sum of the neighbouring concepts of the topic words. Note that we use different weight matrices to distinguish the neighboring concepts in different positions (in head or in tail). This distinction is necessary. For instance, given two knowledge triples \emph{(Big Ben, part of, London)} and \emph{(London, part of, England)}. Even though the concepts \emph{Big Ben} and \emph{England} are both neighboring concepts to \emph{London} with the same relation \emph{part of}, they have the different meaning with regard to \emph{London}. We need to model this difference by \(\mathbf{W}_{\mathbf{3}}\) and \(\mathbf{W}_{\mathbf{4}}\).

Then the final probability of generating a word is computed by 
$$
\mathcal{P}_{t}=\operatorname{softmax}\left(W_{o}\left[d_{t} ; e(s) ; g_{t}\right]+b_{o}\right),
$$ 
where $d_{t}$ is the decoder state at $t$ step and $W_{o} \in \mathbb{R}^{d_{model}\times |V|}$, $b_{o} \in \mathbb{R}^{|V|}$ are trainable decoder parameters. $d_{model}$ is the dimension of \([d_{t} ; e(s) ; g_{t}]\) and $|V|$ is vocabulary size.
\subsection{Topic Label Discriminator}
Another concern is that the generated texts should be closely related to the topic words. To this end, at the second training stage, a topic label discriminator is introduced to perform adversarial training with the SCTKG generator. In a max-min game, the SCTKG generator generates essays to make discriminator consider them semantically match with given topics. Discriminator tries to distinguish the generated essays from real essays. In detail, suppose there are a total of \emph{m} topics, the discriminator produces a sigmoid probability distribution over \(\left(\emph{m} + 1\right)\)classes. The score at \(\left(\emph{m} + 1\right)^{th}\) index represents the probability that the sample is the generated text. The score at the $j^{th}$ \((i \in\{1, \cdots,\emph{m}\})\) index represents the probability that it belongs to the real text with the $j^{th}$ topic. Here the discriminator is a CNN~\cite{Kim2014Convolutional} text classifier.

\subsection{Training}
We introduce our two stage training method in this section. Stage 1: Similar to a conventional CVAE model, The 
loss of our SCTKG generator $-log p(Y |c)$ can be expressed as:
\begin{align}\label{loss}
\begin{aligned}
&-\mathcal{L}\left(\theta ; \phi ; c ; Y \right)_{cvae}=\mathcal{L}_{\mathrm{KL}}+\mathcal{L}_{\mathrm{decoder}}\\
&=\mathrm{KL}\left(q_{\mathrm{\phi}}(z | Y, c) \| p_{\mathrm{\theta}}(z | c)\right)\\
&-\mathbb{E}_{q_{\phi}(z | Y, c)}\left(\log p_{\mathrm{D}}(Y | z, c)\right).
\end{aligned}
\end{align}
Here, $\theta$ and $\phi$ are the parameters of the prior network and recognition network, respectively. Intuitively, $\mathcal{L}_{\mathrm{decoder}}$ maximizes
the sentence generation probability after sampling
from the recognition network, while $\mathcal{L}_{\mathrm{KL}}$
minimizes the distance between the prior and recognition network. Besides, we use the annealing trick and BOW-lossequation \cite{zhao2017learning} to alleviate the vanishing latent variable problem in VAE training.

Stage 2: After trained the SCTKG generator with equation (\ref{loss}), inspired by SeqGan \cite{yu2017seqgan}, we adopt adversarial training between the generator and the topic label discriminator described in section 3.2. We refer reader to \citet{yu2017seqgan} and \citet{yang2019enhancing} for more details.
\section{Experiments}
\subsection{Datasets}\label{4.1}
We conduct experiments on the ZHIHU corpus \cite{feng2018topic}. It consists of Chinese essays\footnote{The dataset can be download by \url{https://pan.baidu.com/s/17pcfWUuQTbcbniT0tBdwFQ}} whose length is between 50 and 100. We select topic words based on frequency and remove rare topic words. The total number of topic labels are set to 100. Sizes of the training set and the test set are 27,000 essays and 2500 essays. For tuning hyperparameters, we set aside 10\% of training samples as the validation set. 

The sentence sentiment labels is required for our model training. To this end, we sample 5000 sentences from the dataset and annotated the data manually with three categories, i.e., positive, negative, neutral. This dataset was divided into a training set, validation set, and test set. We use an open-source Chinese sentiment classifier Senta\footnote{\url{https://github.com/baidu/Senta}} to finetune on our manually-label training set. This classifier achieves an accuracy of 0.83 on the test set. During training, the target sentiment labels $s$ is computed by the sentiment classifier automatically. During inference, users can input any sentiment labels to control the sentiment for sentence generation.

\subsection{Implementation Details}
We use the 200-dim pre-trained word embeddings provided by \citet{song2018directional} and dimension of sentiment embeddings is 32. The vocabulary size is 50,000 and the batch size is 64. We use a manually tuning method to choose the hyperparameter values and the criterion used to select is BLEU \cite{papineni-etal-2002-bleu}. We use GRU with hidden size 512 for both encoder and decoder and the size of latent variables is 300. We implement the model with Tensorflow\footnote{\url{https://github.com/tensorflow/tensorflow}}. The number of parameters is 68M and parameters of our model were randomly initialized over a uniform distribution  [-0.08,0.08]. We pre-train our model for 80 epochs with the MLE method and adversarial training for 30 epochs. The average runtime for our model is 30 hours on a Tesla P40 GPU machine, which adversarial training takes most of the runtime. The optimizer is Adam \cite{kingma2014adam} with $10^{-3}$ learning rate for pre-training and $10^{-5}$ for adversarial training. Besides, we apply dropout on the output layer to avoid over-fitting \cite{srivastava2014dropout} (dropout rate = 0.2) and clip the gradients to the maximum norm of 10. The decoding strategy in this paper uses greedy 
search and average length of generated essays is 79.3.
\subsection{Evaluation}
To comprehensively evaluate the generated essays, we rely on a combination of both automatic evaluation and human evaluation.
\begin{table*}
\centering
\scalebox{0.92} {
\begin{tabular}{l|rrrrr|rrrr}  
\toprule
% &\multicolumn{7}{c}{Automatic Evaluation}& \\
&\multicolumn{5}{c|}{Automatic evaluation}&\multicolumn{4}{c}{Human evaluation}\\
\midrule
Methods  & \textbf{BLEU} & \textbf{Consistency} & \textbf{Novelty} & \textbf{Dist-1} & \textbf{Dist-2} &\textbf{Con.}&\textbf{Nov.}&\textbf{E-div.}&\textbf{Flu.} \\
\midrule
TAV    & 6.05  & 16.59\quad\quad & 70.32& 2.69& 14.25 & 2.32&2.19&2.58&2.76\\
TAT    & 6.32  & 9.19\quad\quad & 68.77& 2.25&  12.17 &1.76 &2.07 &2.32&2.93 \\
MTA    & 7.09  & 25.73\quad\quad & 70.68& 2.24& 11.70  &3.14 & 2.87 & 2.17 &3.25\\
CTEG   & 9.72  & 39.42\quad\quad & 75.71& 5.19& 20.49  &3.74  &3.34 &3.08&3.59\\
\midrule
SCTKG(w/o-Senti) & 9.97 &\textbf{43.84}\quad\quad& 78.32& 5.73&\textbf{23.16}&\textbf{3.89} &3.35 &3.90 &3.71\\
SCTKG(Ran-Senti) &9.64 &41.89\quad\quad&\textbf{79.54}&5.84&23.10&3.80&\textbf{3.48}&\textbf{4.29}&3.67\\
SCTKG(Gold-Senti)  & \textbf{11.02}  & 42.57\quad\quad &78.87 &\textbf{5.92}& 23.07 &  3.81 & 3.37  & 3.94&\textbf{3.75}\\
\bottomrule
\end{tabular}
}
\caption{Automatic and human evaluation result. In human evaluation, \textbf{Con., Nov., E-div., Flu.} represent topic-consistency, novelty, essay-diversity, fluency, respectively. The best performance is highlighted in bold.}
\label{table:1}
\end{table*}

\paragraph{Automatic Evaluation.}
Following previous work \cite{yang2019enhancing}, we consider the following metrics\footnote{\url{https://github.com/libing125/CTEG}}:

\textbf{BLEU}: The BLEU score \cite{papineni2002bleu} is widely used in machine translation, dialogue, and other text generation tasks by measuring word overlapping between ground truth and generated sentences.

\textbf{Dist-1, Dist-2} \cite{Li2015A}: We calculate the proportion of
distinct 1-grams and 2-grams in the generated essays to evaluate the diversity of the outputs.

\textbf{Consistency} \cite{yang2019enhancing}: An ideal essay should closely surround the semantics of all input topics. Therefore, we pre-train a multi-label classifier to evaluate the topic-consistency of the output. A higher “Consistency” score means the generated essays are more closely related to the given topics.
% \textbf{Consistency} is computed by the Jaccard similarity between the output of the classifier and the topic sequence. 

\textbf{Novelty} \cite{yang2019enhancing}: We calculated the novelty by the difference between output and essays with similar topics in the training corpus. A higher “Novelty” score means the output essays are more different from essays in the training corpus.
% We first use the Jaccard similarity function to choose some relevant essays with the input topic in the training set. Then \textbf{Novelty} is computed by the Jaccard similarity between our output essay with the relevant essay chosen above

\textbf{Precision}, \textbf{Recall} and \textbf{Senti-F1}: These metrics are used to measure sentiment control accuracy. If the sentiment label of the generated sentence is consistent with the ground truth, the generated result is right, and wrong otherwise. The sentiment label is predicted by our sentiment classifier mentioned above (see \ref{4.1} for details about this classifier).

\paragraph{Human Evaluation.}
We also perform human evaluation to more accurately evaluate the quality of generated essays. Each item contains the input topics and outputs of different models. Then, 200 items are distributed to 3 annotators, who have no knowledge in advance about the generated essays come from which model. Each annotator scores 200 items and we average the score from three annotators. They are required to score the generated essay from 1 to 5 in terms of three criteria: \textbf{Novelty}, \textbf{Fluency}, and \textbf{Topic-Consistency}. For novelty, we use the TF-IDF features of topic words to retrieve 10 most similar training samples to provide references for the annotators. To demonstrate the paragraph-level diversity of our model, we propose a \textbf{Essay-Diversity} criteria. Specifically, each model generates three essays with the same input topics, and annotators are required to score the diversity by considering the three essays together.
\subsection{Baselines}

\hspace{0.25cm} \textbf{TAV} \cite{feng2018topic} represents topic semantics as the average of all topic embeddings and then uses a LSTM to generate each word. Their work also includes the following two baselines.

\textbf{TAT} \cite{feng2018topic} extends LSTM with an attention mechanism to model the semantic relatedness of each topic word with the generator's output.

\textbf{MTA} \cite{feng2018topic} maintains a topic coverage vector to guarantee that all topic information is expressed during generation through an LSTM decoder.

\textbf{CTEG} \cite{yang2019enhancing} adopts commonsense knowledge and adversarial training to improve generation. It achieves state-of-the-art performance on the topic-to-essay generation task.

\section{Results and Analysis}
In this section, we introduce our experimental results
and analysis from two part: the “text quality”
and “sentiment control”. Then we show case study
of our model.
\subsection{Results on Text Quality}
The automatic and human evaluation results are shown in Table \ref{table:1}. We present three different versions of our model for a comprehensive comparison. (1)``SCTKG(w/o-Senti)" means we do not attach any sentiment label to the model. (2) ``SCTKG(Ran-Senti)'' means we randomly set the sentiment label for each generated sentence. (3) ``SCTKG(Gold-Senti)'' means we set the golden sentiment label for the generated sentence. By investigating the results in Table \ref{table:1}, we have the following observations:

First, all versions of our SCTKG models outperform the baselines in all evaluation metrics (except the BLEU score of SCTKG(Ran-Senti)). This demonstrates that our SCTKG model can generate better essays than baseline models, whether uses the true sentiment, random sentiment or without any sentiment. 

Second, we can learn the superiority of the basic architecture of our model through the comparison between SCTKG(w/o-Senti) and the baselines. In human evaluation, SCTKG(w/o-Senti) outperform CTEG in topic-consistency, essay-diversity, and fluency by +0.15 (3.74 vs 3.89), +0.82 (3.08 vs 3.90), +0.12 (3.59 vs 3.71) respectively. Similar improvements can be also drawn from the automatic evaluation. The improvement in essay-diversity is the most significant. This improvement comes from our CVAE architecture because our sentence representation comes from the sampling from a continuous latent variable. This sampling operation introduces more randomness compared with baselines.

Third, as previously stated, each model generates three essays and considers them as a whole when comparing
the ``E-div”. When given the random and diverse sentiment label sequences, our SCTKG(Ran-Senti) achieves the highest ``E-div'' score (4.29). Consider that CVAE architecture has already improved the diversity compared with baselines. By randomizing the sentiment of each sentence, SCTKG(Ran-Senti) further boosts this improvement (from +0.82 to +1.21 compared with CTEG). This result demonstrates the potential of our model to generate discourse-level diverse essays by using diverse sentiment sequences, proving our claim in the introduction part.

Fourth, when using the golden sentiment label, SCTKG(Gold-Senti) achieves the best performance in BLEU (11.02). However, we find the SCTKG(Gold-Senti) do not significantly outperforms other SCTKG models in other metrics. The results show the true sentiment label of the target sentence benefits SCTKG(Gold-Senti) to better fit in the test set, but there is no obvious help for other important metrics such as diversity and topic-consistency. 

Fifth, we find it interesting that when removing the sentiment label, the SCTKG(w/o-Senti) achieves the best topic-consistency score. We conceive that sentiment label may interfere with the topic information in the latent variable to some extent. But the effect of this interference is trivial. Comparing SCTKG(w/o-Senti) and SCTKG(Gold-Senti), the topic-consistency only drops 0.08 (3.89 vs 3.81) for human evaluation and 1.27 (43.84 vs 42.57) for automatic evaluation, which is completely acceptable for a sentiment controllable model.

%  The improvement in ``Novelty" shows that by encoding extra relation information with concepts bring richer representation to the knowledge base and therefore make the generated text more novel.
\begin{table}
\centering
\scalebox{0.95} {
\resizebox{80mm}{13mm}{
\begin{tabular}{l|rrrccr}  
\toprule
Methods  & \textbf{BLEU} &\textbf{Con.}&\textbf{Nov.}&\textbf{E-div.}&\textbf{Flu.}\\
\midrule
Full model  & \textbf{11.02}  & \textbf{3.81} &\textbf{3.37} &\textbf{3.94}& \textbf{3.75} \\
\midrule
w/o TGA    & 10.34  & 3.54 & 3.17& 3.89&  3.38  \\
w/o AT    & 9.85  & 3.37 & 3.20& 3.92& 3.51 \\
\bottomrule
\end{tabular}}}
\caption{Ablation study on text quality. ``w/o AT'' means without adversarial training. ``w/o TGA'' means withou TGA. \textbf{Con., Nov., E-div., Flu.} represent topic-consistency, novelty, essay-diversity, fluency, respectively. Full model represent SCTKG(Gold-Senti) in this table.}
\label{table:3}
\end{table}

\paragraph{Ablation study on text quality.} To understand how each component of our model contributes to the task, we train two ablated versions of our model: without adversarial training (``w/o AT") and without TGA (``w/o TGA"). Noted that in the ``w/o TGA" experiment, we implement a memory network the same as \citet{yang2019enhancing} which uses the concepts in ConceptNet but regardless of their correlation. All models uses golden sentiment labels. Table \ref{table:3} presents the BLEU scores and human evaluation results of the ablation study.

By comparing full model and “w/o TGA”, we find that without TGA, the model performance drops in all metrics. In particularly, topic-consistency drops 0.27, which shows that by directly learning the correlation between the topic words and its neighboring concepts, concepts that are more closely related to the topic words are given higher attention during generation. Novelty drops 0.2, the reason is that TGA is an expansion of the external knowledge graph information. Therefore the output essays are more novel and informative. Fluency drops 0.37 because TGA benefits our model to choose a more suitable concept in the topic knowledge graph according to the current context. And the BLEU drops for 0.68 shows TGA helps our model to better fit the dataset by modeling the relations between topic words and neighboring concepts. 
% These results show that 
% through complete modeling of the topic knowledge graph, our model utilizes the commonsense knowledge in a more sufficient way and therefore benefits the quality of output texts.

By comparing full model and ``w/o AT", we find that adversarial training can improve the BLEU, topic-consistency, and fluency. The reason is that the discriminative signal enhancing the topic consistency and authenticity of the generated texts.

% \begin{table}
% \centering
% \scalebox{0.95} {
% \begin{tabular}{c|ccc}  
% \toprule
% Methods  & \textbf{Con.} &\textbf{Flu.}&\textbf{M-Div}\\
% \midrule
% Golden Senti    & 3.81  & \textbf{3.75}  & 3.94\\
% Random Senti& 3.80 & 3.67 & \textbf{4.29}\\
% w/o-Senti & \textbf{3.89}&3.71 &3.90\\
% \bottomrule
% \end{tabular}
% }
% \caption{``Golden Senti" means that we use true sentiment label for each sentence, ``Random Senti'' means that we randomly set the sentiment label. ``w/o-Senti'' means do not provide sentiment label. \textbf{Con., M-div., Flu.} represent topic-consistency, multi-diversity, fluency, respectively. }
% \label{table:6}
% \end{table}

\subsection{Results on Sentiment Control}
In this section, we investigate whether the model accurately control the sentiment and how each component affects our sentiment control performance. We train three ablated versions of our model: without sentiment label in encoder, without sentiment label in decoder, and without TGA. We randomly sample 50 essays in our test set with 250 sentences. Instead of using golden sentiment labels, the sentiment labels are randomly given in this section. Predicting the golden sentiment is relatively simple because sometimes emotional labels can be directly derived from the coherence between contexts. We adopt a more difficult experimental setting that aims to generate sentences following arbitrary given sentiment labels. The results are shown in Table \ref{table:5}. 

% To answer question 1, we take human evaluation on \textbf{topic-consistency}, \textbf{fluency}, and \textbf{multi-diversity}, the result can be seen in Table \ref{table:6}. We can learn that our sentiment-attach model has a slight drop in topic consistency, which is consistent with the previous experimental result. The ``Golden Senti'' model performs best in fluency, which means that following the true sentiment transition in the test set leads to a better fluency text. And the ``Random Senti" outperforms the other two models in  the``multi-diversity" metric by a large margin. 
% Note that given a topic sequence, each of the model generate three essays when comparing the ``multi-diversity''. The sentiment label sequences are different in ``Random Senti'' model while remaining identical in ``Golden Senti'' and ``w/o-Senti''. Therefore, the improvement in ``multi-diversity" for ``Random Senti'' demonstrates the potential of our model to generate discourse-level diverse texts by designing diverse sentiment sequences.

\begin{table}
\centering
\scalebox{0.95} {
\begin{tabular}{l|ccc}  
\toprule
Methods  & \textbf{Precison} &\textbf{Recall}&\textbf{Senti-F1}\\
\midrule
Full model  & \textbf{0.68}  & \textbf{0.66} &\textbf{0.67}\\
\midrule
w/o Enc-senti    & 0.56  & 0.55  & 0.56\\
w/o Dec-senti    & 0.59  & 0.62 & 0.61\\
w/o TGA   & 0.62  &0.64  & 0.63\\
\bottomrule
\end{tabular}
}
\caption{Ablation study on sentiment control. ``w/o Enc-senti" means to remove the sentiment embedding in the encoder side and ``w/o Dec-senti" means to remove from the decoder. Full model represents SCTKG(Ran-Senti) in this table.}
\label{table:5}
\end{table}

We can learn that removing the sentiment label either from encoder or decoder leads to an obvious control performance decrease (-11\% / -6\% on Senti-F1) and the sentiment label in the encoder is the most important, since removing it leads to the most obvious decline (-11\% Senti-F1). Although TGA does not directly impose sentiment information, it still helps to improve the control ability (4\% in Senti-F1), which shows that learning correlations among concepts in topic knowledge graph strengthens the emotional control ability of the model. For instance, when given a positive label, the concepts related to the relation ``desire of" are more likely to attach more attention, because the concepts with the relation ``desire of" may represent more positive meaning.

\begin{table}
\centering
\scalebox{0.9} {
\begin{tabular}{l}  
\toprule
\textbf{Input topics}: Law  Education \\
\midrule
\textbf{Sentiment label}: neu. \textcolor{red}{pos.} \textcolor{blue}{neg.} \textcolor{blue}{neg.} neu.\\
%\midrule
\midrule
\textbf{Output essay}: I am a senior high school student.\\ \textcolor{red}{I am in the best high school in our town.} \textcolor{blue}{But} \\\textcolor{blue}{bullying still exist on our campus}. \textcolor{blue}{Teachers} \\\textcolor{blue}{always ignore this phenomenon}. What should we \\ do to protect our rights?\\
\bottomrule
\end{tabular}
}
\caption{Given topic ``Law" and ``Education", and randomly set sentiment label for each sentence. We generated an essay according to the topic and sentiment labels. ``neu.'' represents neutral. ``pos.'' represents positive and ``neg.'' represents negative. We have translated the original Chinese output into English.}
\label{table:4}
\end{table}

\subsection{Case Study}

Table \ref{table:4} presents an example of our output essay with a random sentiment sequence. Positive sentences are shown in red and negative sentences are shown in blue. We can learn that the output essay is not only closely related to the topic ``Law'' and ``Education'', but also corresponding with the randomly given sentiment label. Meanwhile, our model makes full use of commonsense knowledge with the help of TGA. For example, ``high school student" and ``right" are the neighboring concepts related to the topic words ``Education" and ``Law".

% Given the topic ``Love'' ``Experience'' ``Emotion'', and context, our model outputs three sentences following the corresponding sentiment label. In Table \ref{table:4}, we generate three diverse essays with the topic ``Law" ``Education". The results show that our model not only can generate discourse-level diverse texts but also make the most of commonsense knowledge with the help of our TGA method. For example, ``high school student", ``right" in essay 1, ``delinquent" in essay 2 are all the neighbouring concepts and closely related to the topic words ``Law" and ``Education".

% \begin{table}
% \centering
% % \resizebox{\textwidth}{50mm}{
% \begin{tabular}{l}  
% \toprule
% \textbf{Input topics}: Love Experience Emotion\\
% % \specialrule{0.2em}{2pt}{2pt}
% \midrule
% \emph{Context}: It‘s been half a year since I fell
% in love\\ with my boyfriend, he treats me very well.\\
% % \specialrule{0.2em}{2pt}{2pt}
% \midrule
% \emph{positive}: I feel so sweet with him.\\
% \midrule
% \emph{negative}: But he rarely contacts me lately.\\
% \midrule
% \emph{Neural}: We went to a park nearby last week.\\
% \bottomrule
% \end{tabular}
% \caption{Case of sentiment control}
% \label{table:7}
% \end{table}

\section{Related Work}

\paragraph{Topic-to-Text Generation.}
Automatically generating an article is a challenging task in natural language processing. \citet{feng2018topic} are the first to propose the TEG task and they utilize coverage vector to integrate topic information.  \citet{yang2019enhancing} use extra commonsense knowledge to enrich the input information and adopt adversarial training to enhancing topic-consistency. However, both of them fail to consider the sentiment factor in the essay generation and fully utilize the external knowledge base. These limitations hinder them from generating high-quality texts. 

Besides, Chinese poetry generation is similar to our task, which can also be regarded as a topic-to-sequence learning task. \citet{li2018generating} adopt CVAE and adversarial training to generate diverse poetry. \citet{yang2017generating} use CVAE with hybrid decoders to generate Chinese poems. And \citet{yi2018automatic} use reinforcement learning to directly improve the diversity criteria. However, their models are not directly applicable to TEG task. Because they do not take  knowledge into account, their models cannot generate long and meaningful unstructured essays.

\paragraph{Controllable Text Generation.}
Some work has explored style control mechanisms for text generation tasks. For example, \citet{zhou2017mojitalk} use naturally annotated emoji Twitter data for emotional response generation. \citet{wang2018sentigan} propose adversarial training to control the sentiment of the texts. \citet{inproceedings} propose a semi-supervised CVAE to generate poetry and deduce a different lower bound to capture generalized sentiment-related semantics. 
Different from their work, we inject sentiment label in both encoder and decoder of CVAE and prove that by modeling a topic knowledge graph can further enhance the sentiment control ability.

% CVAE \cite{sohn2015learning} extend the conventional VAE model \cite{kingma2013auto} with an additional conditioned label to guide the generation process. Recent research in dialogue generation shows that language generated by VAE models benefit from a significantly greater diversity in comparison with conventional Seq2Seq models \cite{zhao2017learning} and diversity is also a crucial metric in our task, which forms the motivation to use CVAE-based model of our work.

\section{Conclusions}

In this paper, we make a further step in a challenging topic-to-essay generation task by proposing a novel sentiment-controllable topic-to-essay generator with a topic knowledge graph enhanced decoder, named SCTKG. To get better representation from external knowledge, we present TGA, a novel topic knowledge graph representation mechanism. Experiments show that our model can not only generate sentiment-controllable essays but also outperform competitive baselines in text quality.

\bibliography{anthology,emnlp2020}
\bibliographystyle{acl_natbib}

\end{document}